\documentclass[sigconf, screen, nonacm]{acmart}

\usepackage{threeparttable}
\usepackage{pifont}
\usepackage{multirow}
\usepackage{multicol}
\usepackage{bbding}
\usepackage{makecell, multirow}
\usepackage{amsfonts}
\usepackage{float}
\usepackage{bbding}
\usepackage{subfigure}
\usepackage{xcolor}
\usepackage{color}
\usepackage{makecell}
\usepackage{colortbl}
\AtBeginDocument{%
  }

\begin{document}

\title{Focus on What You Need: Global-to-Local Dynamic Attention for Multimodal Sarcasm Detection}
\author{Dehao Wu}
\authornote{Both authors contributed equally to this research.}
\affiliation{%
  \institution{University of Illinois Urbana-Champaign}
  \country{}
}
\email{dehaowu2@illinois.edu}

\author{Xingjie Zhuang}
\authornotemark[1]
\affiliation{%
  \institution{Xiamen University}
  \country{}
}
\email{zhuangxj@xmu.stu.edu.cn}

\renewcommand{\shortauthors}{Dehao Wu and Xingjie Zhuang}

\begin{abstract}
Multimodal sarcasm detection has attracted increasing attention due to the widespread use of images and text on social media platforms. Existing studies generally treat the semantic or emotional inconsistency between images and text as a key cue for identifying sarcastic expressions. However, images in real-world scenarios often contain multiple visual objects, many of which are irrelevant to sarcasm-related semantics. Such redundant visual information may introduce noise during cross-modal interaction, distracting models from focusing on critical sarcasm cues and potentially leading to inaccurate predictions. To address this issue, we propose \textbf{GLDA} (\textbf{G}lobal-to-\textbf{L}ocal \textbf{D}ynamic \textbf{A}ttention), a multimodal sarcasm detection framework that models cross-modal interactions from a global-to-local perspective. Specifically, GLDA first employs a global multimodal fusion module to capture high-level semantic interactions between visual and textual modalities, enabling the model to identify coarse-grained cross-modal inconsistencies. To further reduce the influence of irrelevant objects, GLDA introduces a local gated cross-modal attention mechanism that dynamically filters redundant visual and textual features and refines multimodal representations at a finer granularity. In addition, a multi-branch learning objective is designed to jointly optimize unimodal and multimodal representations. Experimental results demonstrate that GLDA effectively improves sarcasm detection performance and achieves competitive results compared with existing approaches. The code will be available at 
\href{https://anonymous.4open.science/r/GLDA-DBE6}{\textcolor{red!75}{https://anonymous.4open.science/r/GLDA}}.
\end{abstract}



\maketitle
\section{Introduction}
Sarcasm, a ubiquitous form of figurative language, operates on the delicate tension between literal expression and intended sentiment. Given its profound impact on sentiment analysis and public opinion monitoring, the task of sarcasm detection has evolved into a pivotal research frontier \cite{ref1,ref2}. While early methodologies focused exclusively on linguistic cues \cite{ref3,ref4}, the surge of multimedia content has rendered text-only analysis insufficient. In the absence of visual or acoustic context, the subtle "ironic gap" in user-generated content often remains elusive, driving the field toward multimodal sarcasm detection (MSD) \cite{ref5,ref6,ref7,ref8}.

Contemporary MSD research has shifted from simple feature aggregation to sophisticated cross-modal interaction. For instance, researchers have leveraged interactive graph convolutional networks to map intricate intra-modal and inter-modal inconsistencies \cite{ref9}, or harnessed multi-granularity cues within the multi-view framework to capture nuanced sarcastic signals \cite{ref8}. Furthermore, by integrating external knowledge, recent models have begun to simulate the hierarchical semantic coherence between text and imagery \cite{ref10}. Beyond architectural innovation, a new paradigm is emerging that addresses model robustness and fairness. To combat inherent dataset flaws, recent studies have introduced causal-based debiasing frameworks to strip away statistical label biases \cite{ref7}. Others have pioneered counterfactual data augmentation, utilizing adaptive contrastive learning to decouple sarcastic intent from biased textual factors \cite{ref11}, thereby ensuring more generalized and objective detection
\begin{figure}[t!]   
	\centering
 \includegraphics[width=1\linewidth]{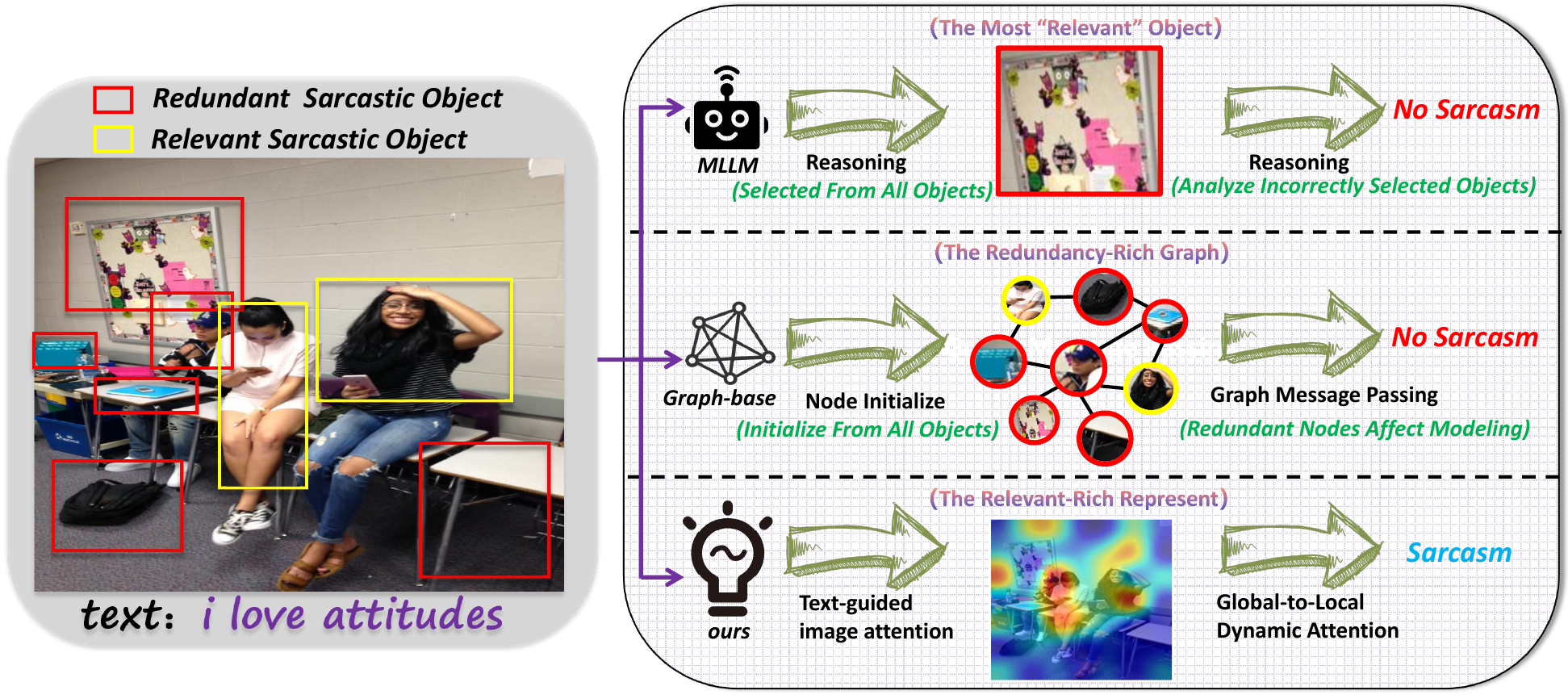}
	\caption{Comparison of our method with the other exist multimodal methods on the multimodal sarcasm detection task.}
	\label{Figure1}
\end{figure}

Based on the aforementioned studies, it is not difficult to observe that a general consensus has been formed in the field of multimodal sarcasm detection: most researchers regard the semantic or emotional inconsistency between images and text as an important cue for understanding sarcastic expressions.

However, in practical sarcasm examples, the input image often contains multiple visual objects. It is important to note that not all objects in an image are strongly related to the sarcastic cues expressed in the corresponding text. In many cases, a large portion of the visual content is essentially redundant information. Such redundant visual signals not only fail to assist the model in capturing the key sarcasm cues, but may also distract the model from focusing on truly relevant elements, thereby weakening the effectiveness of cross-modal reasoning. Even recent multimodal large language models are not immune to this problem when dealing with images that contain substantial visual distractions, as illustrated in Figure \ref{Figure1}. This is largely due to architectural characteristics and context-length constraints of multimodal large language models, such as Gemini, which prevent them from simultaneously reasoning over all visual objects in complex scenes. As a result, these models tend to adaptively focus on a limited number of objects that they consider most relevant to the textual context during inference. Nonetheless, when redundant objects are present in the image, they may interfere with the model’s selection process and lead the model to attend to suboptimal or irrelevant visual cues, thereby affecting the overall reasoning quality. This issue becomes particularly pronounced in approaches that rely on graph-based structures or region-level interaction modeling, which are widely regarded as mainstream paradigms for modeling fine-grained cross-modal interactions, where a large number of irrelevant nodes can introduce significant noise and hinder the modeling of meaningful cross-modal relationships. 

Notably, to alleviate this issue, some recent studies attempt to introduce additional visual features extracted by object detection models such as Faster R-CNN and YOLO, aiming to obtain more fine-grained region-level representations for subsequent cross-modal interaction. Although this strategy may reduce the risk of missing potentially important visual objects during feature extraction, it essentially increases the number of candidate regions rather than addressing the core challenge of filtering irrelevant information. Consequently, the additional object-level features may still contain a large amount of noise that is unrelated to sarcasm understanding. Moreover, incorporating extra detection models inevitably introduces additional computational overhead and increases the overall model complexity.

To tackle these challenges, this paper proposes a multimodal sarcasm detection framework that models cross-modal interactions from a global-to-local perspective and adopts a multi-branch collaborative learning strategy. Specifically, the model first employs a global multimodal fusion module to capture the overall semantic interactions between visual and textual modalities, enabling the model to identify high-level cross-modal inconsistencies. Subsequently, a local gated cross-modal attention mechanism is introduced to refine the information flow at a finer granularity. Through a gating mechanism, the model can dynamically filter irrelevant visual objects and textual tokens, thereby focusing on sarcasm-related semantic cues while suppressing redundant information. In addition, a multi-branch training strategy is adopted to jointly optimize unimodal and multimodal representations, which helps the model learn robust modality-specific features while maintaining effective cross-modal interaction.

By modeling multimodal interactions in a global-to-local manner, the proposed framework is able to capture both coarse-grained semantic inconsistencies and fine-grained sarcasm cues between images and text, thereby improving the overall performance of multimodal sarcasm detection. The main contributions of this work are summarized as follows:
\begin{itemize} 
    \item We investigate the impact of redundant visual objects in multimodal sarcasm detection, revealing that irrelevant visual cues can disrupt cross-modal interaction and hinder the capture of critical sarcasm signals.
    
    \item We propose a global-to-local multimodal framework that combines global semantic inconsistency modeling with a local gated cross-modal attention mechanism to refine representations and suppress redundant visual information.
    
    \item We introduce a multi-branch optimization strategy to jointly enhance unimodal robustness and cross-modal interaction, leading to consistent improvements on benchmark datasets.
\end{itemize}

\section{Related Work} 

\subsection{Multimodal Sarcasm Detection} 
Traditional sarcasm detection primarily focuses on identifying sarcastic expressions by analyzing the semantic relationships among words or sentences within textual contexts \cite{ref12, ref17,ref18}. However, with the rapid proliferation of social media platforms, user-generated content increasingly combines multiple modalities such as images and text. This trend has driven growing interest in multimodal sarcasm detection, which aims to understand sarcastic intent by jointly modeling information from different modalities.

Early studies laid the foundation for this task. \cite{ref13} first formally defined the problem of multimodal sarcasm detection, while \cite{ref14} introduced a public dataset containing multimodal sarcastic content and proposed a hierarchical fusion framework for sarcasm recognition. Building upon these works, subsequent studies explored various cross-modal interaction mechanisms. For instance, \cite{ref9,ref15,ref16} employed graph neural networks to model semantic relationships across modalities from multiple perspectives, enabling models to capture both intra-modal and cross-modal inconsistencies. Meanwhile, \cite{ref8} constructed multiple multimodal views, allowing the model to capture sarcasm-related cues at different semantic granularities. More recently, researchers have also investigated the impact of data bias in multimodal sarcasm detection. For example, \cite{ref11} proposed an adaptive debiasing contrastive learning strategy to mitigate statistical label bias and non-sarcastic word bias, while \cite{ref7} introduced a causal graph framework to model variable relationships and reduce bias through counterfactual reasoning

In contrast to these approaches that mainly rely on complex cross-modal interaction architectures, Global-to-Local Dynamic Attention (GLDA) framework focuses on dynamically identifying sarcasm-relevant multimodal cues while suppressing redundant visual information.
\subsection{Cross-modal Alignment}
Cross-modal alignment is generally analyzed via coarse-grained and fine-grained perspectives. Coarse-grained methods often rely on global metrics like AC Score\cite{ref19} or Modality Integration Rate (MIR) \cite{ref21}, which effectively evaluate overall pre-training but struggle to capture token-level details. For fine-grained alignment, while Logit Lens Analysis \cite{ref22} identifies object-specific tokens, it lacks explicit optimization. Other works use QA-based coordination \cite{ref23} or predefined vocabularies with contrastive loss (e.g., SEA \cite{ref24}), though the latter is limited by fixed categories. Recent improvements \cite{ref25} leverage automatic annotation from RAM \cite{ref26} and efficient cosine similarity to bypass vocabulary constraints.

To achieve precise local alignment, vision grounding models like GLIP \cite{ref28} and Grounding DINO \cite{ref27} learn directly from region-text pairs, while RegionCLIP \cite{ref25} uses pseudo-labels. However, high annotation costs remain a bottleneck. To mitigate this, self-distillation methods (e.g., CLIPSelf \cite{ref30}, MaskEmbed \cite{ref31}) align region features with [CLS] tokens or use reconstruction tasks, though they are prone to teacher model bias and foreground interference. Lastly, audio-visual alignment utilizes attention mechanisms (e.g., TPAVI \cite{ref32}) for spatiotemporal synchronization. While filtering silent regions via masks \cite{ref33} reduces noise, such rigid strategies may hinder the model's ability to distinguish between active and silent objects, limiting complex multimodal interaction modeling.

In contrast to previous cross-modal alignment methods that primarily aim to explicitly match visual regions with textual tokens, our proposed framework focuses on dynamically selecting sarcasm-relevant multimodal cues and suppressing redundant visual objects, enabling more effective cross-modal reasoning for multimodal sarcasm detection.
\section{Methodology} 
\subsection{Problem Formulation}
Given an image--text pair $(I, T)$, where $I$ denotes an input image and $T = \{w_1, w_2, ..., w_L\}$ represents a textual sequence containing $L$ tokens, the goal of multimodal sarcasm detection is to predict whether the pair expresses sarcasm. The corresponding label is defined as $y \in \{0,1\}$.

In real-world social media scenarios, images usually contain multiple visual objects, while texts consist of many words. However, only a subset of these objects and words are directly related to sarcasm expression. The remaining elements typically constitute redundant information that may introduce noise and mislead the model during cross-modal interaction. Such redundancy can distract the model from focusing on key sarcasm cues and weaken its capability of identifying semantic incongruity between modalities.

To address this challenge, we propose a multimodal sarcasm detection framework that progressively filters redundant information and highlights critical semantic cues. The framework consists of three major components:
\begin{itemize}
\item \textbf{Unimodal Representation Learning}
\item \textbf{Cross-modal Interaction Learning}
\item \textbf{Multi-branch Optimization Criteria}
\end{itemize}

\begin{figure*}[t!]
    \centering
 \includegraphics[width=0.9\linewidth]{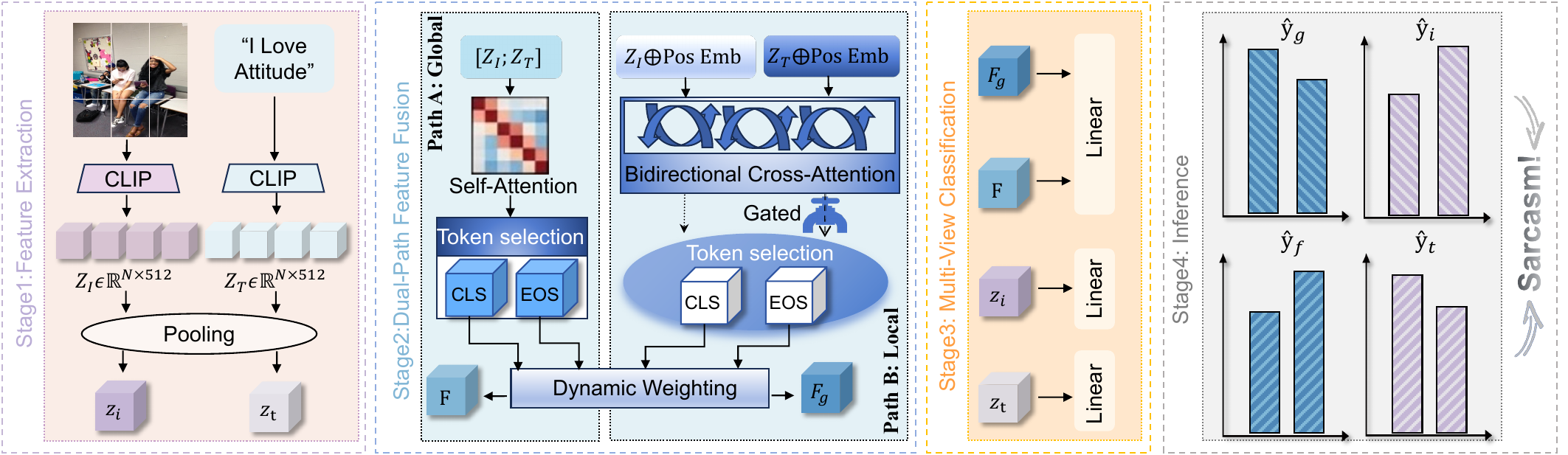}
    \caption{The overall architecture of the proposed GLDA framework. In Stage~1, a CLIP encoder extracts image and text token representations $Z_I$ and $Z_T$, which are pooled into global vectors $\mathbf{z}_i$ (image) and $\mathbf{z}_t$ (text). In Stage~2, Path~A (Global) fuses $Z_I$ and $Z_T$ via self-attention to produce the global representation $\mathbf{F}$, while Path~B (Local) applies bidirectional cross-attention with a gating mechanism to produce the gated representation $F_g$. In Stage~3, three linear classifiers are applied: a shared classifier for $\mathbf{F}$ and $F_g$, and two modality-specific classifiers for $\mathbf{z}_i$ and $\mathbf{z}_t$, yielding predictions $\hat{y}_\text{f}$, $\hat{y}_\text{g}$, $\hat{y}_\text{i}$, and $\hat{y}_\text{t}$. These predictions are aggregated in Stage~4 for final sarcasm prediction.}
    \label{Figure2}
\end{figure*}

The overall architecture of the proposed framework is illustrated in Figure~\ref{Figure2}.

\subsection{Unimodal Representation Learning}

Given an image--text pair $(I,T)$, our goal is to learn reliable 
modality-specific representations before performing cross-modal reasoning. 
We first employ a pretrained vision--language encoder to independently 
extract visual and textual features.

For the textual modality, the text encoder transforms the input sequence 
$T=\{w_1,w_2,\dots,w_L\}$ into token-level representations:
\begin{equation}
\mathbf{H}_T = \mathrm{Enc}_T(T), \quad 
\mathbf{H}_T \in \mathbb{R}^{L \times d},
\end{equation}
where $L$ denotes the number of tokens and $d$ is the embedding dimension. Similarly, the visual encoder extracts patch-level representations from the input image:
\begin{equation}
\mathbf{H}_I = \mathrm{Enc}_I(I), \quad 
\mathbf{H}_I \in \mathbb{R}^{N \times d},
\end{equation}
where $N$ represents the number of visual patches. To align the two modalities into a shared embedding space, we apply 
learnable projection layers:
\begin{equation}
\mathbf{Z}_T = \mathbf{W}_T \mathbf{H}_T, \qquad
\mathbf{Z}_I = \mathbf{W}_I \mathbf{H}_I,
\end{equation}
where $\mathbf{W}_T$ and $\mathbf{W}_I$ denote projection matrices. Global modality representations are obtained via pooling:
\begin{equation}
\mathbf{z}_t = \mathrm{Pool}(\mathbf{Z}_T), \qquad
\mathbf{z}_i = \mathrm{Pool}(\mathbf{Z}_I),
\end{equation}

These representations are used to generate unimodal predictions:
\begin{equation}
\hat{y}_t = \mathbf{W}_t \mathbf{z}_t + b_t, \qquad
\hat{y}_i = \mathbf{W}_i \mathbf{z}_i + b_i,
\end{equation}
where $\hat{y}_t$ and $\hat{y}_i$ denote the predicted sarcasm probabilities 
from the textual and visual branches, respectively.

\subsection{Cross-modal Interaction Learning}
Sarcasm often emerges from semantic incongruity between visual and textual modalities. However, both images and texts typically contain redundant information. Images may include multiple objects unrelated to sarcasm, while textual descriptions may contain tokens that do not contribute to sarcastic meaning. Such redundancy can mislead cross-modal interaction and dilute the model's attention to relevant cues. To mitigate this, we develop a dual-granularity interaction mechanism: 1) a Global Multimodal Fusion module to capture holistic dependencies for comprehensive context understanding; 2) a Local Gated Cross-modal Attention module that acts as a precision filter. The latter assigns gating weights to individual elements, suppressing noise to distill the subtle, localized incongruities essential for detection.


\subsubsection{Global Multimodal Fusion Module}
To capture global cross-modal relationships while preserving modality-specific semantics, we design a global multimodal fusion module based on a Transformer encoder.

We first concatenate visual and textual representations to construct a joint multimodal sequence:
\begin{equation}
\mathbf{H} = [\mathbf{Z}_I ; \mathbf{Z}_T], 
\end{equation}
where $\mathbf{Z}_I \in \mathbb{R}^{N \times d}$ and $\mathbf{Z}_T \in \mathbb{R}^{L \times d}$ denote the projected visual and textual representations, respectively. Here $N$ represents the number of visual patches, $L$ denotes the number of textual tokens, and $d$ is the embedding dimension. $\mathbf{H} \in \mathbb{R}^{(N+L) \times d}$ therefore forms a unified multimodal sequence that integrates both visual and textual features.

A Transformer encoder is then applied to model global cross-modal interactions:
\begin{equation}
\mathbf{H}_f = \mathrm{Transformer}(\mathbf{H}),
\end{equation}
where $\mathbf{H}_f$ denotes the fused multimodal representations produced by the Transformer encoder after modeling interactions between visual and textual elements.

After interaction, modality-specific representations are obtained via pooling:
\begin{equation}
\mathbf{h}_i = \mathrm{Pool}(\mathbf{H}_f^{img}), \qquad
\mathbf{h}_t = \mathrm{Pool}(\mathbf{H}_f^{txt}),
\end{equation}
where $\mathbf{H}_f^{img}$ and $\mathbf{H}_f^{txt}$ denote the visual and textual parts of the fused feature sequence, respectively. The pooling operation aggregates token-level representations into global modality-level vectors $\mathbf{h}_i$ and $\mathbf{h}_t$.

Since different modalities may contribute unequally to sarcasm detection, we introduce a dynamic weighting mechanism:
\begin{equation}
(w_t, w_i) = 
\mathrm{Softmax}\left(
\mathbf{W}_a \mathbf{h}_t , 
\mathbf{W}_a \mathbf{h}_i
\right),
\end{equation}
where $\mathbf{W}_a$ is a learnable parameter matrix used to estimate the importance of each modality. The Softmax function normalizes the modality scores into importance weights $w_t$ and $w_i$, representing the relative contributions of textual and visual features.

The final fused representation is computed as:
\begin{equation}
\mathbf{F} = w_t \mathbf{h}_t + w_i \mathbf{h}_i,
\end{equation}
where $\mathbf{F}$ denotes the final multimodal representation that dynamically integrates textual and visual information and is used for subsequent sarcasm classification.

\subsubsection{Local Gated Cross-modal Attention}

While global fusion captures high-level multimodal interactions, it may inadvertently incorporate noise from irrelevant objects or tokens.

\begin{figure}[t!]   
	\centering
 \includegraphics[width=0.9\linewidth]{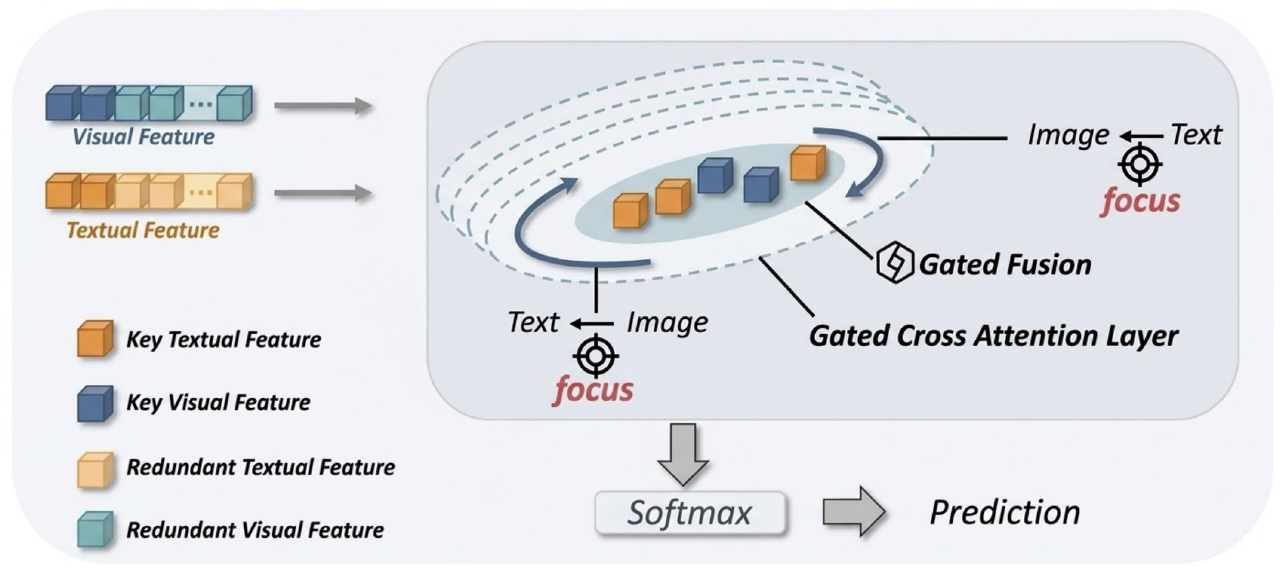}
	\caption{Local gated cross-modal attention schematic diagram.}
	\label{Figure3}
\end{figure}

To address this issue and enhance the selection of relevant information, we introduce a local gated cross-modal attention mechanism to refine the information flow at a finer granularity. By focusing on local-level correlations, this gated module acts as a selective filter to suppress redundant features, as shown in Figure~\ref{Figure3}. Specifically, we first perform bidirectional cross-modal attention:
\begin{equation}
\mathbf{H}_I' =
\mathrm{Attn}(\mathbf{Z}_I, \mathbf{Z}_T, \mathbf{Z}_T),
\end{equation}
\begin{equation}
\mathbf{H}_T' =
\mathrm{Attn}(\mathbf{Z}_T, \mathbf{Z}_I, \mathbf{Z}_I),
\end{equation}
where $\mathbf{Z}_I$ and $\mathbf{Z}_T$ denote the projected visual and textual representations introduced in the previous module. 
$\mathbf{H}_I'$ and $\mathbf{H}_T'$ represent the updated visual and textual features after cross-modal attention. The attention function is defined as:
\[
\mathrm{Attn}(\mathbf{Q}, \mathbf{K}, \mathbf{V})
=
\mathrm{Softmax}
\left(
\frac{\mathbf{Q}\mathbf{K}^\top}{\sqrt{d}}
\right)
\mathbf{V},
\]
where $\mathbf{Q}$, $\mathbf{K}$, and $\mathbf{V}$ denote the query, key, and value matrices, respectively, and $d$ represents the feature embedding dimension used for scaling.

To control the amount of cross-modal information integrated into the original representations, a gating mechanism is introduced:
\begin{equation}
\mathbf{G} =
\sigma
\left(
\mathbf{W}_g
[\mathbf{H}_T' ; \mathbf{Z}_T]
\right),
\end{equation}
where $\mathbf{W}_g$ denotes a learnable parameter matrix and $\sigma(\cdot)$ represents the sigmoid activation function. The gate vector $\mathbf{G}$ determines how much cross-modal information should be incorporated into the original textual representation.

The updated textual representation is computed as:
\begin{equation}
\tilde{\mathbf{H}}_T
=
\mathbf{G} \odot \mathbf{H}_T'
+
(1-\mathbf{G}) \odot \mathbf{Z}_T,
\end{equation}
where $\odot$ denotes element-wise multiplication. 
$\tilde{\mathbf{H}}_T$ represents the refined textual feature that adaptively integrates the original representation and the cross-modal interaction.

A similar operation is applied to obtain the refined visual representation $\tilde{\mathbf{H}}_I$. After pooling these gated features, we obtain modality-level representations $\mathbf{h}_t^{g}$ and $\mathbf{h}_i^{g}$.

The gated multimodal representation is then computed as:
\begin{equation}
\mathbf{F}_g =
w_t \mathbf{h}_t^{g}
+
w_i \mathbf{h}_i^{g},
\end{equation}
where $w_t$ and $w_i$ are the modality importance weights introduced in the global fusion module. 
$\mathbf{F}_g$ denotes the final gated multimodal representation used for subsequent sarcasm prediction.

\subsection{Multi-branch Optimization Criteria}

To effectively train the proposed framework, we adopt a multi-branch 
optimization criteria that primarily optimizes cross-modal interaction representations while introducing auxiliary supervision for unimodal 
branches. This design is consistent with the architecture of our model, 
which consists of a global multimodal fusion module, a local gated 
cross-modal attention module, and unimodal prediction branches. By jointly supervising these components, the model is encouraged to capture the semantic inconsistency between image and text—one of the key characteristics of sarcasm—while maintaining strong modality-specific representations.

\subsubsection{Global Fusion Loss.}
The primary cross-modal supervision is defined on the fused multimodal representation produced by the global fusion module:
\begin{equation}
L_{\mathrm{fuse}} =
\mathrm{CE}(\hat{y}_f , y),
\end{equation}
where $\mathrm{CE}(\cdot)$ denotes the cross-entropy loss, $\hat{y}_f$ represents the prediction generated from the fused multimodal representation, and $y$ is the ground-truth sarcasm label. This loss encourages the model to learn global cross-modal representations that capture interactions between visual and textual modalities.

\subsubsection{Local Gated Interaction Loss.}
To further enhance fine-grained cross-modal reasoning, we impose an 
additional supervision signal on the local gated cross-modal attention 
branch:
\begin{equation}
L_{\mathrm{gated}} =
\mathrm{CE}(\hat{y}_g , y),
\end{equation}

This objective guides the gating mechanism to filter redundant visual 
objects and irrelevant textual tokens, allowing the model to focus on 
sarcasm-related semantic cues at a finer granularity.

\subsubsection{Unimodal Auxiliary Losses.}
Although sarcasm detection mainly relies on cross-modal interaction, 
individual modalities may still contain useful semantic signals. Therefore, we introduce auxiliary supervision for the unimodal prediction branches:
\begin{equation}
L_{\mathrm{text}} =
\mathrm{CE}(\hat{y}_t , y), \qquad
L_{\mathrm{image}} =
\mathrm{CE}(\hat{y}_i , y),
\end{equation}
where $\hat{y}_t$ and $\hat{y}_i$ denote predictions derived from the textual and visual representations, respectively. These auxiliary objectives encourage each modality to learn informative features independently and help stabilize multimodal training.

Finally, the overall training objective is defined as the weighted 
combination of the above losses:
\begin{equation}
L =
L_{\mathrm{fuse}}
+
L_{\mathrm{gated}}
+
\lambda_1 L_{\mathrm{text}}
+
\lambda_2 L_{\mathrm{image}}.
\end{equation}
where $\lambda_1$ and $\lambda_2$ control the contribution of unimodal supervision during training. By jointly optimizing these objectives, the model learns robust unimodal representations while effectively capturing both global multimodal interactions and fine-grained cross-modal relationships for sarcasm detection.

\section{Experimental Setup} 
\subsection{\textbf{Dataset}}
For multimodal sarcasm detection, we conduct experiments on MMSD and MMSD 2.0 \cite{ref8}. MMSD \cite{ref14} consists of English image-text pairs from Twitter, where labels are automatically assigned based on sarcasm-related hashtags. MMSD 2.0 \cite{ref8} improves upon MMSD by removing misleading cues and re-annotating noisy samples, providing a more reliable benchmark. Both datasets are split into training, validation, and test sets with an 80\%/10\%/10\% ratio.

For multimodal fake news detection, we evaluate on the Weibo \cite{ref34} and Twitter \cite{ref35} datasets. The Weibo dataset contains 13,272 Chinese image-text pairs labeled as rumor or non-rumor, while the Twitter dataset includes 15,991 image-text pairs collected from real-world events and labeled as fake or real. Both datasets follow the same 80\%/10\%/10\% data split.
The basic statistics of these datasets are summarized in Tables \ref{table:1}.
\begin{table}[htbp]
  \centering
  \caption{Statistics of datasets for multi-modal sarcasm detection and rumor detection.}
  \resizebox{1\linewidth}{!}{
    \begin{tabular}{c|c|c|cc|cc|cc}
    \toprule
    \multirow{2}[3]{*}{Task} & \multirow{2}[3]{*}{Datasets} & \multirow{2}[3]{*}{Total} & \multicolumn{2}{c|}{Train} & \multicolumn{2}{c|}{Val} & \multicolumn{2}{c}{Test} \\
    \cmidrule{4-9}          
                            &                             &                           & \#Positive & \#Negative & \#Positive & \#Negative & \#Positive & \#Negative \\
    \midrule
    \multirow{2}[0]{*}{Sarcasm Detection} 
                            & MMSD\cite{ref14}                       & 19816                     & 8642       & 11174      & 959        & 1451       & 959        & 1450       \\
                            & MMSD2.0\cite{ref8}                     & 19816                     & 9576       & 10240      & 1042       & 1368       & 1037       & 1372       \\
    \midrule
    \multirow{2}[0]{*}{Rumor Detection}
                            & Weibo\cite{ref34}                        & 13272                     & 6263       & 4290       & 864       & 528        & 827       & 500       \\
                            & Twitter\cite{ref35}                      & 15991                     & 7518       & 6187       & 608       & 621       & 660        & 397        \\
    \bottomrule
    \end{tabular}%
    }
  \label{table:1}%
\end{table}%

\subsection{Experimental Settings}
GLDA is implemented based on CLIP (ViT-B/32) as the pre-trained vision–language backbone. The visual encoder extracts $N{=}50$ patch tokens, while the text encoder processes sequences of up to $L{=}77$ tokens. Both modalities are projected into a shared embedding space of dimension $d{=}512$ for cross-modal interaction. The model is trained end-to-end using the multi-branch objective in Eq.~(19).

We use AdamW as the optimizer with a batch size of 128 and train for 10 epochs. The learning rate is set to $5\times10^{-7}$ for the CLIP backbone and $1\times10^{-4}$ for the cross-modal modules, with a warmup proportion of 0.1 and weight decay of 0.01. The maximum gradient norm is clipped to 5, and a dropout rate of 0.3 is applied. The hidden size is set to 512 with 8 attention heads, and the number of gated cross-modal layers is 3. The unimodal loss weights are $\lambda_1{=}0.8$ and $\lambda_2{=}0.7$. All hyperparameters are tuned on the validation set. Experiments are conducted on a single NVIDIA A6000 48G GPU.


\subsection{Comparison Methods}
\subsubsection{Single-modal methods.}
SMSD \cite{ref18} is a self-matching network using low-rank bilinear pooling. BERT \cite{ref36}, LLaMA3.1-8B \cite{ref37}, and Mixtral \cite{ref40} provide bidirectional, general-purpose, and sparse MoE-based textual representations, respectively. For visual modeling, ResNet \cite{ref38} and ViT \cite{ref39} extract features via pooling layers and CLS tokens to support multimodal detection.

\subsubsection{Multi-modal methods.}
HFM \cite{ref14} uses attribute-initialized Bi-LSTM for feature fusion. D\&R Net \cite{ref41} captures cross-modal contrastive relationships. InCrossMGs \cite{ref15} and CMGCN \cite{ref9} employ GCNs to learn modal interactions, with the latter using attribute-object pairs. Multi-View CLIP \cite{ref8} integrates multi-granularity cues, while DIP \cite{ref6} applies contrastive reweighting for factual-emotional analysis. MoBa \cite{ref42} optimizes parameter efficiency. ESAM \cite{ref45} decouples features with sentimental constraints. ITFNet \cite{ref43} models tone-polarization via tension fields, and LLaVA-v1.5 \cite{ref44} combines CLIP and Vicuna for instruction-based reasoning.

\section{Experimental Results and Analysis}
\subsection{Performance Comparison}
The effectiveness of our method is validated through extensive comparisons with strong baselines (Table \ref{table:2}). First, GLDA consistently outperforms the strongest unimodal baselines, highlighting the necessity of multimodal modeling for sarcasm detection. This is largely due to the intrinsic contradiction between literal expression and intended meaning in sarcasm, which is difficult to capture from a single modality but can be effectively modeled through cross-modal cues.

Second, large language models (e.g., LLaMA3.1-8B and Mistral-7B) perform poorly in zero-shot settings, indicating limited capability in understanding sarcasm without task-specific fine-tuning. Notably, the multimodal model LLaVA-v1.5-7B underperforms even text-only LLMs in zero-shot scenarios, revealing the challenge of directly applying general-purpose multimodal models to this task. Only after supervised fine-tuning does LLaVA achieve competitive performance, even surpassing strong proprietary models such as GPT-4V.

\begin{table}[htp]
  \centering
  \caption{Comparison of different methods on MMSD \cite{ref14} and MMSD2.0 \cite{ref8} dataset, where LoRA denotes the model fine-tuned with LoRA-based supervised fine-tuning (SFT).
  }
  \resizebox{1\linewidth}{!}{
    \begin{tabular}{c|cccc|cccc}
    \toprule
    \multirow{2}[4]{*}{\textbf{Method}} & \multicolumn{4}{c|}{\textbf{MMSD}} & \multicolumn{4}{c}{\textbf{MMSD2.0}} \\
    \cmidrule{2-9}          
    & Acc.(\%) & P(\%) & R(\%) & F1(\%) & Acc.(\%) & P(\%) & R(\%) & F1(\%) \\
    \midrule
    \multicolumn{9}{c}{\textit{\textbf{Image-Only Methods}}} \\
    \midrule
    ResNet\cite{ref38} & 64.76  & 54.41  & 70.80  & 61.53  & 65.50  & 61.17  & 54.39  & 57.58  \\
    ViT\cite{ref39}   & 67.83  & 57.93  & 70.07  & 63.40  & 72.02  & 65.26  & 74.83  & 69.72  \\
    \midrule
    \multicolumn{9}{c}{\textit{\textbf{Text-Only Methods}}} \\
    \midrule
    SMSD\cite{ref18}   & 80.90  & 76.46  & 75.18  & 75.82  & 73.56  & 68.45  & 71.55  & 69.97  \\
    BERT\cite{ref36}   & 83.60  & 78.50  & 82.51  & 80.45  & 76.52  & 74.48  & 73.09  & 73.91  \\
    \cellcolor{gray!20}LLaMA3.1-8B (Zero-Shot) \cite{ref37} & \cellcolor{gray!20}60.58  & \cellcolor{gray!20}58.48  & \cellcolor{gray!20}59.97  & \cellcolor{gray!20}59.21  & \cellcolor{gray!20}58.52 & \cellcolor{gray!20}56.15  & \cellcolor{gray!20}57.46  & \cellcolor{gray!20}56.79  \\
    \cellcolor{gray!20}Mistral-7B (Zero-Shot) \cite{ref40} & \cellcolor{gray!20}60.15  & \cellcolor{gray!20}57.99  & \cellcolor{gray!20}58.42  & \cellcolor{gray!20}58.20  & \cellcolor{gray!20}57.81 & \cellcolor{gray!20}55.86  & \cellcolor{gray!20}56.31  & \cellcolor{gray!20}56.03  \\
    \cmidrule{1-9} 
    \multicolumn{9}{c}{\textit{\textbf{Multi-Modal Methods}}} \\
    \midrule
    HFM\cite{ref14}   & 83.44  & 76.57  & 84.15  & 80.18  & 70.57  & 64.84  & 69.05  & 66.88  \\
    D\&R Net\cite{ref41} & 84.02  & 77.97  & 83.42  & 80.60  & -     & -     & -     & - \\
    InCrossMGs\cite{ref15} & 86.10  & 81.38  & 84.36  & 82.84  & -     & -     & -     & - \\
    CMGCN\cite{ref9} & 86.54  & -     & -     & 82.73  & 79.83  & 75.82  & 78.01  & 76.90  \\
    Multi-View CLIP\cite{ref8} & 88.33  & 82.66  & 88.65  & 85.55  & 85.64  & 80.33  & 88.24  & 84.10  \\
    DIP\cite{ref6}    & 89.59 & 87.76  & 86.58  & 87.17  & 85.32  & 81.76  & 85.94  & 83.79  \\
    MoBA\cite{ref42}  & 88.96  & 82.84  & 88.12  & 85.40  & 85.83  & 80.42  & 88.67  & 84.34  \\
    ESAM\cite{ref45}  & 90.11  & 86.87  & 89.54  & 88.19  & 85.87  & 83.12  & 86.05  & 84.56  \\
    ITFNet\cite{ref43} & 92.04& 90.21  & 90.30 & 90.25  & 86.73 & 81.08  & 88.03  & 84.41  \\
    \cellcolor{gray!20}GPT-4V (Zero-Shot)\cite{ref46} & \cellcolor{gray!20}71.04  & \cellcolor{gray!20}59.43  & \cellcolor{gray!20}96.52  & \cellcolor{gray!20}73.56  & \cellcolor{gray!20}-  & \cellcolor{gray!20}-  & \cellcolor{gray!20}-  & \cellcolor{gray!20}-  \\
    \cellcolor{gray!20}LLaVA-v1.5-7B (Zero-Shot)\cite{ref44} & \cellcolor{gray!20}-  & \cellcolor{gray!20}-  & \cellcolor{gray!20}-  & \cellcolor{gray!20}-  & \cellcolor{gray!20}47.92  & \cellcolor{gray!20}44.16  & \cellcolor{gray!20}78.64  & \cellcolor{gray!20}56.53  \\
    \cellcolor{gray!20}LLaVA-v1.5-7B (LoRA)\cite{ref44} & \cellcolor{gray!20}93.67  & \cellcolor{gray!20}93.70  & \cellcolor{gray!20}93.14  & \cellcolor{gray!20}93.40  & \cellcolor{gray!20}85.18  & \cellcolor{gray!20}85.59  & \cellcolor{gray!20}85.20  & \cellcolor{gray!20}85.11  \\
    \cmidrule{1-9}          
    \cellcolor{cyan!20}$\textbf{GLDA}$ & \cellcolor{cyan!20}\textbf{92.31} & \cellcolor{cyan!20}\textbf{87.83}& \cellcolor{cyan!20}\textbf{94.18} & \cellcolor{cyan!20}\textbf{90.91} & \cellcolor{cyan!20}\textbf{86.45} & \cellcolor{cyan!20}\textbf{81.61} & \cellcolor{cyan!20}\textbf{89.64} & \cellcolor{cyan!20}\textbf{85.44} \\
    \bottomrule
    \end{tabular}%
  }
  \label{table:2}%
\end{table}%

Finally, compared with classical multimodal approaches, GLDA achieves consistent improvements across datasets, validating its effectiveness. Although slightly inferior to SFT-enhanced LLaVA on certain metrics, GLDA demonstrates clear advantages in efficiency, requiring substantially less training time and approximately half the GPU memory (Table \ref{table:3}). Overall, it achieves a favorable balance between performance and computational cost.
\begin{table}[htbp]
  \centering
  \caption{Training time and GPU memory usage of different models.}
    \resizebox{0.95\linewidth}{!}{
  \begin{tabular}{lccc}
    \toprule
    \textbf{Model} & \textbf{Training time(Hour)} & \textbf{GPU memory(GB)} \\
    \midrule
    LLaVA-v1.5-7B (LoRA)   & 12.3 & 39.2 \\
    \textbf{GLDA }           & \textbf{0.6} & \textbf{19.8} \\
    \bottomrule
  \end{tabular}
  }
  \label{table:3}
\end{table}

\subsection{Ablation Studies}
To evaluate the contribution of each component in GLDA, we conduct ablation studies from both branch-level and interaction-level perspectives (Table \ref{table:4}).

\subsubsection{Branch-based ablation.} Removing any branch degrades performance, confirming all components are beneficial. In particular, removing the global cross-modal branch causes the most significant drop, indicating that high-level semantic inconsistency modeling is the dominant factor in sarcasm detection. Meanwhile, removing the local gated branch also leads to noticeable degradation, demonstrating its role in refining representations by filtering redundant information. These results suggest that the global and local branches are complementary, with the former capturing coarse cross-modal alignment and the latter enabling fine-grained refinement.

\subsubsection{Interaction-based ablation.} We further analyze the role of cross-modal interaction by comparing models with and without interaction. Results show that models without cross-modal interaction perform significantly worse, highlighting the necessity of modeling cross-modal inconsistency. Moreover, combining global and local interactions yields further gains, validating the effectiveness of multi-level cross-modal interaction.
\begin{table}[htbp]
  \centering
  \caption{The ablation results of modules in the model. $\mathcal{B}_{t}$, $\mathcal{B}_{i}$, $\mathcal{B}_{l}$ and $\mathcal{B}_{g}$ denote the text-only modality learning branch, the image-only modality learning branch, the global cross-modal learning branch, and the local cross-modal learning branch, respectively.}
  \resizebox{0.85\linewidth}{!}{
    \begin{tabular}{cccc|cc|cc}
    \toprule
    \multirow{2}[3]{*}{$\mathcal{B}_{t}$} & \multirow{2}[3]{*}{$\mathcal{B}_{i}$} & \multirow{2}[3]{*}{$\mathcal{B}_{l}$} & \multirow{2}[3]{*}{$\mathcal{B}_{g}$} & \multicolumn{2}{c|}{MMSD} & \multicolumn{2}{c}{MMSD2.0} \\
    \cmidrule{5-8}          
    &       &       &       & Acc.(\%) & F1(\%) & Acc.(\%) & F1(\%) \\
    \midrule
    \multicolumn{8}{l}{\textbf{Category 1: Branch-based Ablation}} \\
    \midrule
    & \Checkmark & \Checkmark & \Checkmark & 90.15 & 89.23 & 85.06 & 84.75 \\ 
    \Checkmark & & \Checkmark & \Checkmark & 90.58 & 89.67 & 85.44 & 84.96 \\ 
    \Checkmark & \Checkmark & & \Checkmark & 89.42 & 88.15 & 84.68 & 83.23 \\ 
    \Checkmark & \Checkmark & \Checkmark & & 90.41 & 89.12 & 85.31 & 84.39 \\ 
    \midrule
    \multicolumn{8}{l}{\textbf{Category 2: Modality-based Ablation}} \\
    \midrule
    \Checkmark & \Checkmark & & & 87.86 & 86.95 & 82.02 & 81.70 \\ 
    & & \Checkmark & \Checkmark & 89.98 & 89.01 & 84.92 & 84.26 \\
    \midrule
    \multicolumn{8}{l}{\textbf{Category 3: Full Model}} \\
    \midrule
    \Checkmark & \Checkmark & \Checkmark & \Checkmark & \textbf{92.31} & \textbf{90.91} & \textbf{86.45} & \textbf{85.44} \\
    \bottomrule
    \end{tabular}%
    }
  \label{table:4}%
\end{table}%

\subsection{Hyperparameter Analysis}
To evaluate the sensitivity of GLDA to key hyperparameters, we conduct all analyses on MMSD2.0, whose reduced noise provides a more reliable setting. Considering the inherent class imbalance, we adopt F1 score as the primary metric.

\begin{figure}[t!]   
	\centering
 \includegraphics[width=1\linewidth]{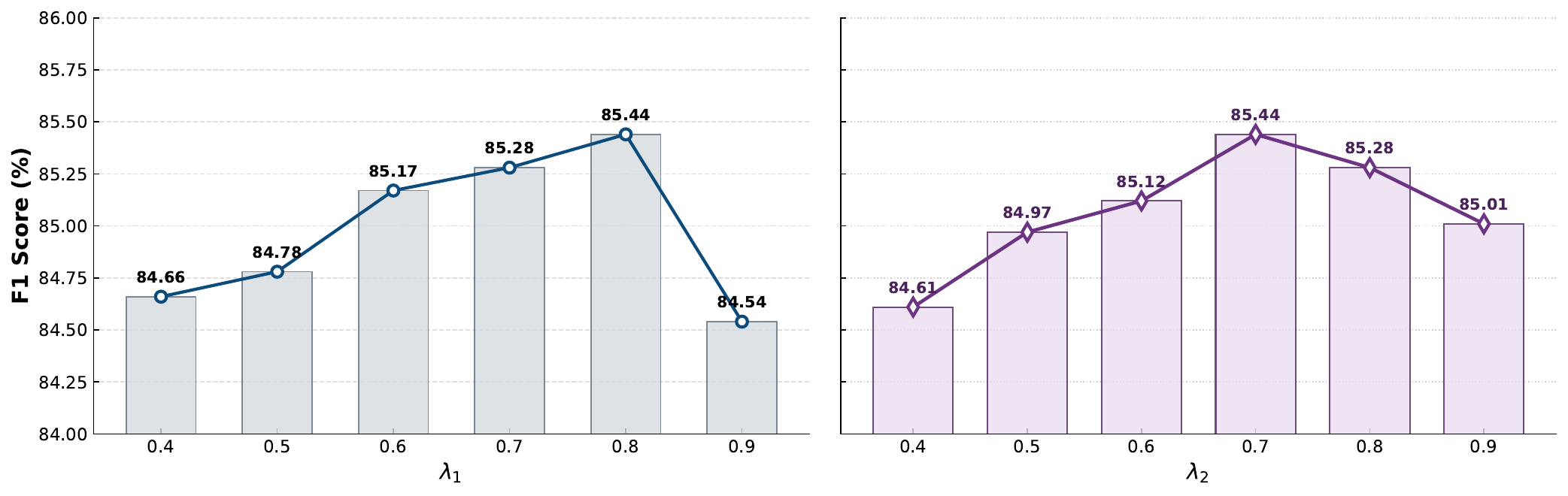}
	\caption{Parameter sensitivity analysis of $\lambda_1$ and $\lambda_2$.}
	\label{Figure4}
\end{figure}

We first analyze the unimodal loss weights $\lambda_1$ and $\lambda_2$ in the multi-branch objective (Figure~\ref{Figure4}). Both exhibit a bell-shaped trend, with performance peaking at $\lambda_1=0.8$ and $\lambda_2=0.7$. This indicates that moderate unimodal supervision improves representation learning, while excessive weighting disrupts the balance between unimodal learning and cross-modal fusion. Notably, the optimal visual weight is lower than the textual one, reflecting that visual inputs often contain redundant objects. A relatively smaller $\lambda_2$ prevents overfitting to noisy visual signals, while the gated mechanism further mitigates sensitivity to visual supervision.

\begin{figure}[t!]   
	\centering
 \includegraphics[width=0.8\linewidth]{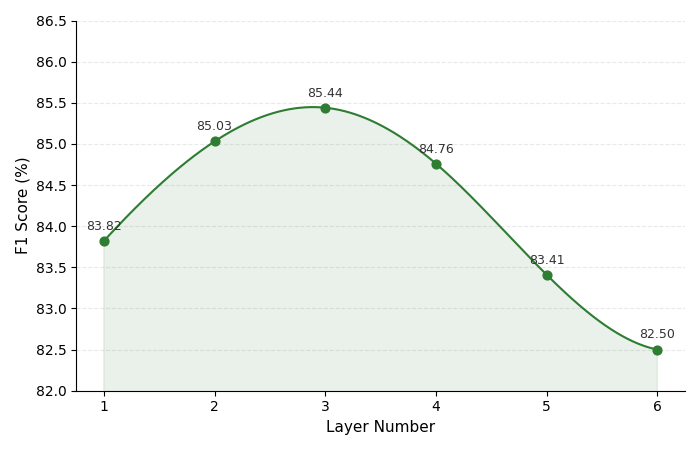}
	\caption{Parameter sensitivity analysis of gated cross-attention layers.}
	\label{Figure5}
\end{figure}

We further vary the number of gated cross-attention layers, as summarized in Figure~\ref{Figure5}. Performance increases with depth and peaks at three layers, then degrades with additional layers, forming an inverted-U pattern. This suggests a trade-off in local cross-modal interaction: insufficient depth limits fine-grained alignment, while excessive depth leads to over-smoothing, weakening discriminative cross-modal contrasts. These findings highlight that effective sarcasm detection requires controlled, rather than excessive, cross-modal interaction.


\subsection{Generalization Ability Analysis}
Whether a method can be effectively generalized to other tasks has always been a highly concerned issue. To explore this, we extend evaluation to multimodal rumor detection, a related task that also relies on identifying cross-modal inconsistency. While sarcasm involves sentiment incongruity between text and image, rumor detection often exhibits mismatches between visual evidence and textual claims. This shared property provides a natural testbed for evaluating the generality of GLDA. Notably, we retrain the same architecture on Twitter and Weibo rumor detection datasets and compare against several strong baselines. No structural changes are made, and only dataset-specific hyperparameters are tuned. The Weibo dataset further evaluates cross-lingual generalization due to its Chinese content. Results are reported in Table \ref{table:5}.

GLDA achieves accuracies of 92.7\% and 93.6\% on Twitter and Weibo, respectively, outperforming all baselines. This strong transfer performance stems from two key factors. First, the gated cross-modal attention mechanism models token-level interactions without task-specific bias, enabling it to capture diverse forms of cross-modal inconsistency while filtering irrelevant visual content. Second, the multi-branch objective provides complementary supervision that improves robustness across tasks. The strong performance on Weibo further suggests that GLDA is robust to language variation, as cross-modal reasoning primarily operates in aligned embedding spaces. Overall, these results indicate that modeling cross-modal incongruity provides a general and transferable foundation for multimodal understanding.
\begin{table}[htbp]
  \centering
  \caption{Performance comparison on the Weibo and Twitter datasets.}
  \resizebox{0.9\linewidth}{!}{
    \begin{tabular}{c|c|ccc|ccc}
    \toprule
    \multirow{2}[3]{*}{Model} & \multirow{2}[3]{*}{Acc} & \multicolumn{3}{c|}{Rumor} & \multicolumn{3}{c}{Non-Rumor} \\
    \cmidrule{3-8}
                              &                         & P & R & F1 & P & R & F1 \\
    \midrule
    \multicolumn{8}{c}{\textbf{Weibo Dataset}} \\
    \midrule
    CAFE~\cite{ref47}                       & 0.840 & 0.855 & 0.830 & 0.842 & 0.825 & 0.851 & 0.837 \\
    MRML~\cite{ref48}                       & 0.897 & 0.898 & 0.887 & 0.892 & 0.896 & 0.905 & 0.901 \\
    HMCAN~\cite{ref49}                      & 0.885 & 0.920 & 0.845 & 0.881 & 0.856 & 0.926 & 0.890 \\
    MFIR~\cite{ref50}                       & 0.925 & 0.901 & 0.840 & 0.869 & 0.906 & 0.848 & 0.876 \\
    MIGCL~\cite{ref51}                      & 0.932 & 0.925 & 0.899 & 0.912 & 0.936 & 0.954 & 0.944 \\
    \rowcolor{cyan!20} \textbf{GLDA (Ours)} & \textbf{0.936} & \textbf{0.936} & \textbf{0.961} & \textbf{0.949} & \textbf{0.935} & \textbf{0.893} & \textbf{0.913} \\
    \midrule
    \multicolumn{8}{c}{\textbf{Twitter Dataset}} \\
    \midrule
    CAFE~\cite{ref47}                       & 0.806 & 0.807 & 0.799 & 0.803 & 0.805 & 0.813 & 0.809 \\
    MRML~\cite{ref48}                       & 0.803 & 0.821 & 0.844 & 0.832 & 0.896 & 0.777 & 0.747 \\
    HMCAN~\cite{ref49}                      & 0.897 & 0.971 & 0.801 & 0.878 & 0.853 & 0.979 & 0.912 \\
    MFIR~\cite{ref50}                       & 0.858 & 0.850 & 0.848 & 0.849 & 0.867 & 0.871 & 0.869 \\
    MIGCL~\cite{ref51}                      & 0.906 & 0.886 & 0.932 & 0.908 & 0.929 & 0.881 & 0.904 \\
    \rowcolor{cyan!20} \textbf{GLDA (Ours)} & \textbf{0.927} & \textbf{0.935} & \textbf{0.915} & \textbf{0.925} & \textbf{0.920} & \textbf{0.939} & \textbf{0.929} \\
    \bottomrule
    \end{tabular}
  }
  \label{table:5}
\end{table}

\subsection{Case Study}

We investigate three research questions: (RQ1) How does GLDA identify sarcasm-relevant cues at both the image and token levels? (RQ2) Does cross-modal inconsistency modeling generalize to other tasks? (RQ3) What are the current limitations of the framework?

\subsubsection{RQ1: Sarcasm Cue Identification}

\begin{figure}[t!]
\centering
\includegraphics[width=0.9\linewidth]{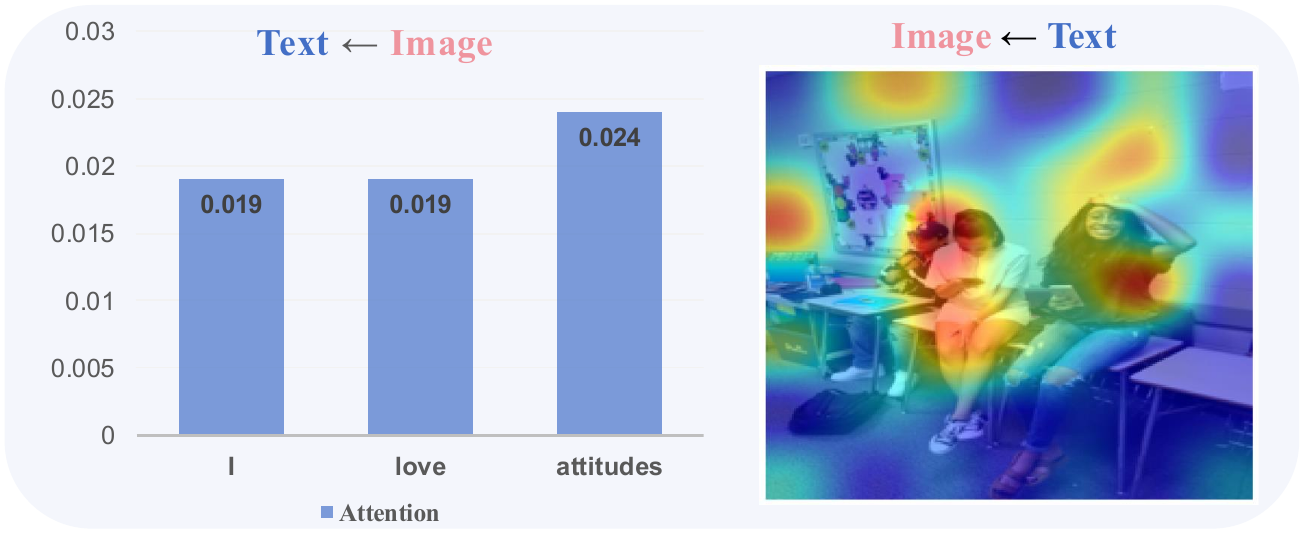}
\caption{Cross-modal attention visualization for a sarcastic sample from MMSD2.0. Left: Grad-CAM heatmap (Image$\leftarrow$Text). Right: gated cross-attention weights (Text$\leftarrow$Image).}
\label{Figure6}
\end{figure}

As shown in Figure~\ref{Figure6}, the text \emph{``I love attitudes''} is paired with an image of a person sitting on a desk in a classroom, using a phone instead of attending the lecture. In the Grad-CAM heatmap, high-activation regions are concentrated on the person and the phone, while background elements such as walls and distant desks are suppressed, indicating that the model attends selectively to content that contradicts the text. In the cross-attention weights, the token \emph{``attitudes''} receives the highest weight, as the gating mechanism identifies it as the key token whose situated meaning---determined by the depicted behavior---directly contradicts the positive framing of \emph{``love.''} These results confirm that GLDA filters irrelevant visual regions at the image level and pinpoints inconsistency-anchoring words at the token level.

\subsubsection{RQ2: Cross-task Generalization}

\begin{figure}[t!]
\centering
\includegraphics[width=\linewidth]{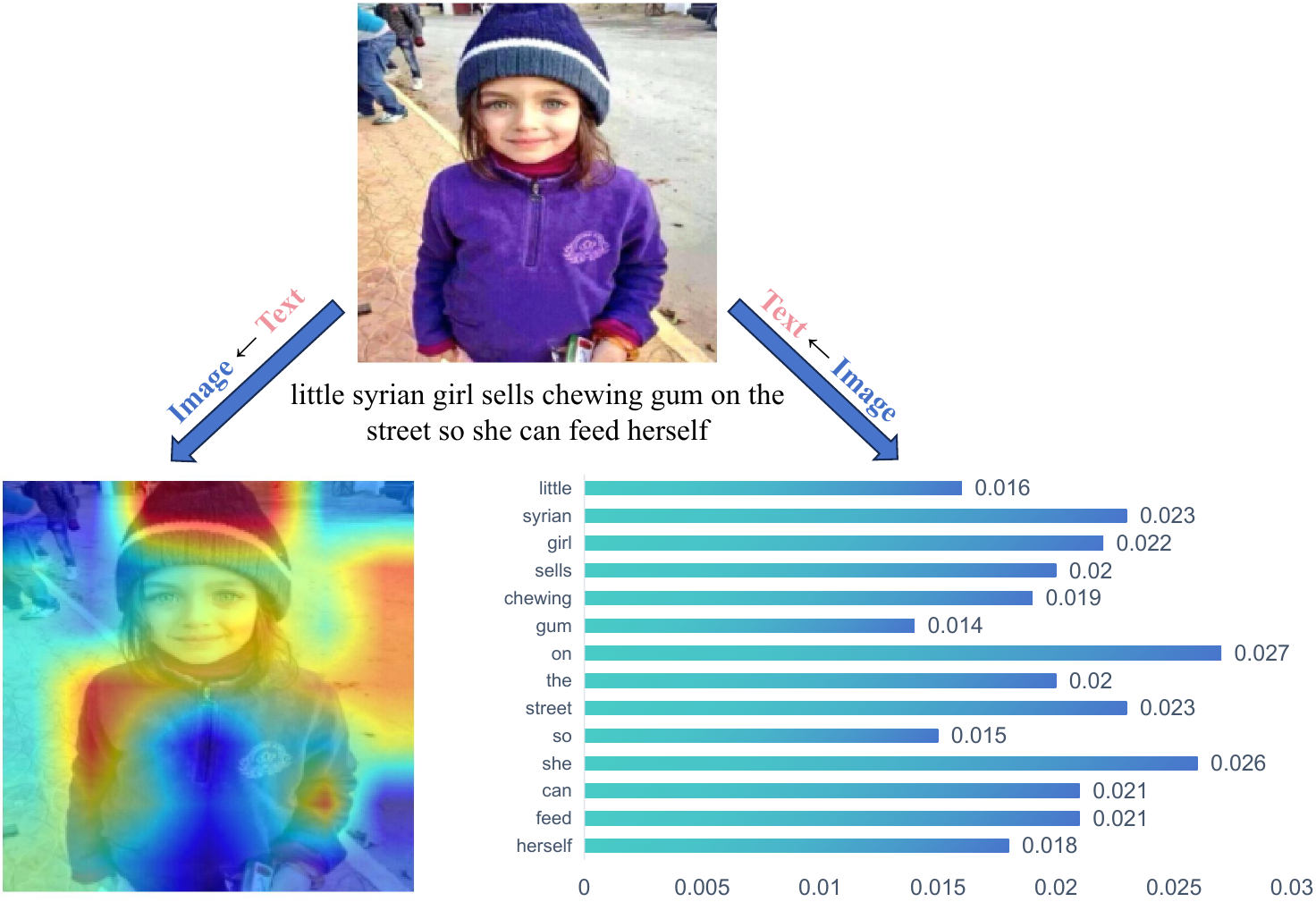}
\caption{Cross-modal attention visualization for a rumor sample from the Twitter dataset. Left: Grad-CAM heatmap (Image$\leftarrow$Text). Right: gated cross-attention weights (Text$\leftarrow$Image).}
\label{Figure7}
\end{figure}

To evaluate the transferability of the learned cross-modal interaction, we apply the same model to a rumor detection sample from the Twitter dataset without any architectural modification. As shown in Figure~\ref{Figure7}, the text \emph{``Little Syrian girl sells chewing gum on the street so she can feed herself''} is paired with an image of a young girl wearing a clean purple fleece jacket and a winter hat, smiling at the camera. The Grad-CAM heatmap highlights the clean, well-maintained clothing rather than the face or background, focusing on visual evidence that contradicts the hardship narrative. Meanwhile, the cross-attention weights are distributed across multiple tokens, reflecting that the inconsistency originates from the overall narrative rather than from a single keyword. This pattern mirrors the mechanism observed in RQ1---detecting semantic discrepancies between textual claims and visual evidence---and demonstrates the ability of the architecture to transfer across tasks without modification.

\subsubsection{RQ3: Limitations}
\begin{figure}[htp]
\centering
\includegraphics[width=0.7\linewidth]{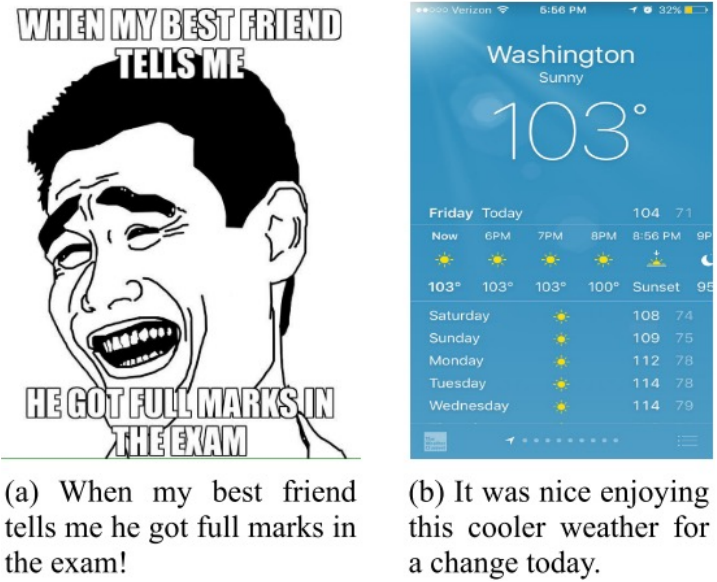}
\caption{Two representative failure cases from MMSD2.0. (a) A meme-based sarcastic sample. (b) A weather-based sarcastic sample.}
\label{Figure8}
\end{figure}

We identify two representative failure patterns from misclassified samples, as illustrated in Figure~\ref{Figure8}. In Figure~\ref{Figure8}(a), the sarcasm relies on a widely recognized meme face whose meaning derives from cultural convention rather than low-level visual attributes, which the visual encoder is unable to capture. In Figure~\ref{Figure8}(b), recognizing the sarcasm requires interpreting that 103\textdegree F denotes extreme heat---a character-level numerical reasoning step that lies beyond the scope of the current vision-language alignment. These cases reveal a clear boundary: GLDA effectively captures inconsistencies grounded in visual appearance, but struggles when comprehension requires external cultural knowledge or symbolic numerical reasoning.

\section{Conclusion and Future Work} 
This paper revisits multimodal sarcasm detection from the perspective of information redundancy and selective cross-modal reasoning. We show that irrelevant visual content fundamentally hinders effective interaction, and argue that what to filter is as important as how to fuse. To this end, we propose GLDA, a global-to-local dynamic attention framework that captures coarse-grained semantic inconsistency and progressively refines representations via gated fine-grained filtering, enabling the model to focus on sarcasm-relevant cues while suppressing noise. Extensive experiments demonstrate that GLDA achieves strong performance with improved efficiency. Ablation and transfer studies consistently highlight a key insight: modeling cross-modal inconsistency under controlled information flow is crucial for robust sarcasm understanding. Moreover, its effectiveness on multimodal rumor detection suggests that such modeling generalizes beyond task-specific patterns.

Despite these advances, challenges remain in handling sarcasm requiring external knowledge (e.g., cultural symbols) and fine-grained symbolic reasoning (e.g., numerical inconsistencies). Future work will explore knowledge-enhanced reasoning and improved fine-grained perception to address these limitations, moving toward more generalizable multimodal understanding frameworks.

\bibliographystyle{ACM-Reference-Format}
\bibliography{references}

\end{document}